\documentclass[preprint,12pt]{elsarticle}

\usepackage{lineno}
\modulolinenumbers[5]

\usepackage[dvipsnames]{xcolor}
\usepackage{multirow}

\usepackage{bbm}
\usepackage{amsmath}
\usepackage{amssymb}

\usepackage{latexsym}
\usepackage{algpseudocode}
\usepackage{blindtext}
\usepackage{algorithm}
\usepackage{times}
\usepackage{epsfig}
\usepackage{graphicx}
\usepackage{textcase}
\usepackage{comment}
\usepackage{breqn}
\usepackage{ragged2e}
\usepackage[normalem]{ulem}
\usepackage{caption, booktabs}

\usepackage{upgreek}
\usepackage{array}
\usepackage{textcomp}

\newcolumntype{P}[1]{>{\centering\arraybackslash}p{#1}}
\useunder{\uline}{\ul}{}
\usepackage{balance}
\usepackage{lastpage}
\usepackage[pagebackref=true,breaklinks=true,letterpaper=true,colorlinks,bookmarks=false]{hyperref}

\usepackage{todonotes}
\usepackage{xparse}
\usepackage{array}
\newcolumntype{L}[1]{>{\raggedright\let\newline\\\arraybackslash\hspace{0pt}}m{#1}}
\newcolumntype{C}[1]{>{\centering\let\newline\\\arraybackslash\hspace{0pt}}m{#1}}
\newcolumntype{R}[1]{>{\raggedleft\let\newline\\\arraybackslash\hspace{0pt}}m{#1}}
\newcolumntype{J}[1]{>{\justifying\let\newline\\\arraybackslash\hspace{0pt}}m{#1}}

\usepackage{enumitem}
\usepackage{float}

\usepackage[capitalize]{cleveref}
\crefname{section}{Sec.}{Secs.}
\Crefname{section}{Section}{Sections}
\Crefname{table}{Table}{Tables}
\crefname{table}{Tab.}{Tabs.}

\newcommand{\srcDomain}{$\mathcal{S}$}
\newcommand{\srcImg}{$\mathit{I}_{s}$}
\newcommand{\trgImg}{$\mathit{I_{t}}$}
\newcommand{\srcLb}{$\mathit{Y}_{s}$}
\newcommand{\trgLb}{$\mathit{Y}_{t}$}

\newcommand{\trgDomain}{$\mathcal{T}$}
\newcommand{\myEq}[1]{Eq.~(#1)}
\DeclareMathOperator*{\argmax}{arg\,max} 

\newcommand{\myComment}[1]{}
\newcolumntype{P}[1]{>{\centering\arraybackslash}p{#1}}

\begin{document}
\begin{frontmatter}

\title{Leveraging Topology for Domain Adaptive Road Segmentation in Satellite and Aerial Imagery}


\author{Javed~Iqbal\corref{cor}}
\cortext[cor]{Corresponding author}
\ead{javed.iqbal@itu.edu.pk}

\author{Aliza~Masood}
\ead{aliza.masood@itu.edu.pk}

\author{Waqas~Sultani }
\ead{waqas.sultani@itu.edu.pk}

\author{Mohsen~Ali}
\ead{mohsen.ali@itu.edu.pk}

\address{Information Technology University, Pakistan}





\begin{abstract}

Getting precise aspects of road through segmentation from remote sensing imagery is useful for many real-world applications such as autonomous vehicles, urban development and planning, and achieving sustainable development goals (SDGs) \footnote{https://sdgs.un.org/goals}.  
Roads are only a small part of the image, and their appearance, type, width, elevation, directions, etc. exhibit large variations across geographical areas.
Furthermore, due to differences in urbanization styles, planning, and the natural environments; regions along the roads vary significantly.  
Due to these variations among the train and test domains (domain shift), the road segmentation algorithms fail to generalize to new geographical locations. 
Unlike the generic domain alignment scenarios, road segmentation has no scene structure and generic domain adaptive segmentation methods are unable to enforce topological properties like continuity, connectivity, smoothness, etc., thus resulting in degraded domain alignment. In this work, we propose a topology-aware unsupervised domain adaptation approach for road \textit{segmentation} in remote sensing imagery.
During domain adaptation for road segmentation, we predict road skeleton, an auxiliary task to enforce the topological constraints.
To enforce consistent predictions of road and skeleton, especially in the unlabeled target domain, the \textit{conformity loss} is defined across the skeleton prediction head and the road-segmentation head. 
Furthermore, for self-training, we filter out the noisy pseudo-labels by using a connectivity-based pseudo-labels refinement strategy, on both road and skeleton segmentation heads, thus avoiding holes and discontinuities. 
Extensive experiments on the benchmark datasets show the effectiveness of the proposed approach compared to existing state-of-the-art methods.
Specifically, for SpaceNet to DeepGlobe adaptation,  the proposed approach outperforms the competing methods by a minimum margin of 6.6\%, 6.7\%, and 9.8\% in IoU, F1-score, and APLS, respectively. 
\end{abstract}


\end{frontmatter}

\section{Introduction}
\label{sec:intro}
Roads, defined simply as \textit{`a wide way leading from one place to another'} \footnote{https://www.lexico.com/definition/road}, play a vital role in limiting or expanding the opportunities to connect.
Building a reliable,  up-to-date, and comprehensive map of roads with attributes like width, number of lanes, type, etc., is crucial not only for economic analysis, public policy, future development (building smart cities), and private businesses but to the modern world of autonomous transportation too. 
However, segmenting the unlabeled roads or updating the labels and attributes for the existing ones is labor-intensive,  manually cumbersome, and costly \cite{herfort2021evolution}, which becomes further crucial and time-sensitive in the case of natural disasters.
Some recent works used graph-based approaches to extract and map road networks \cite{he2020sat2graph,tan2020vecroad}. However, these graph-based approaches only predict the center lines as edges between two possible nodes and are not suitable for extracting road characteristics like road width, number of lanes,  type, etc. In this work, we specifically focus on the segmentation and adaptation of roads in remote-sensing imagery.

\begin{figure*}[t!]
    \includegraphics[width=\linewidth]{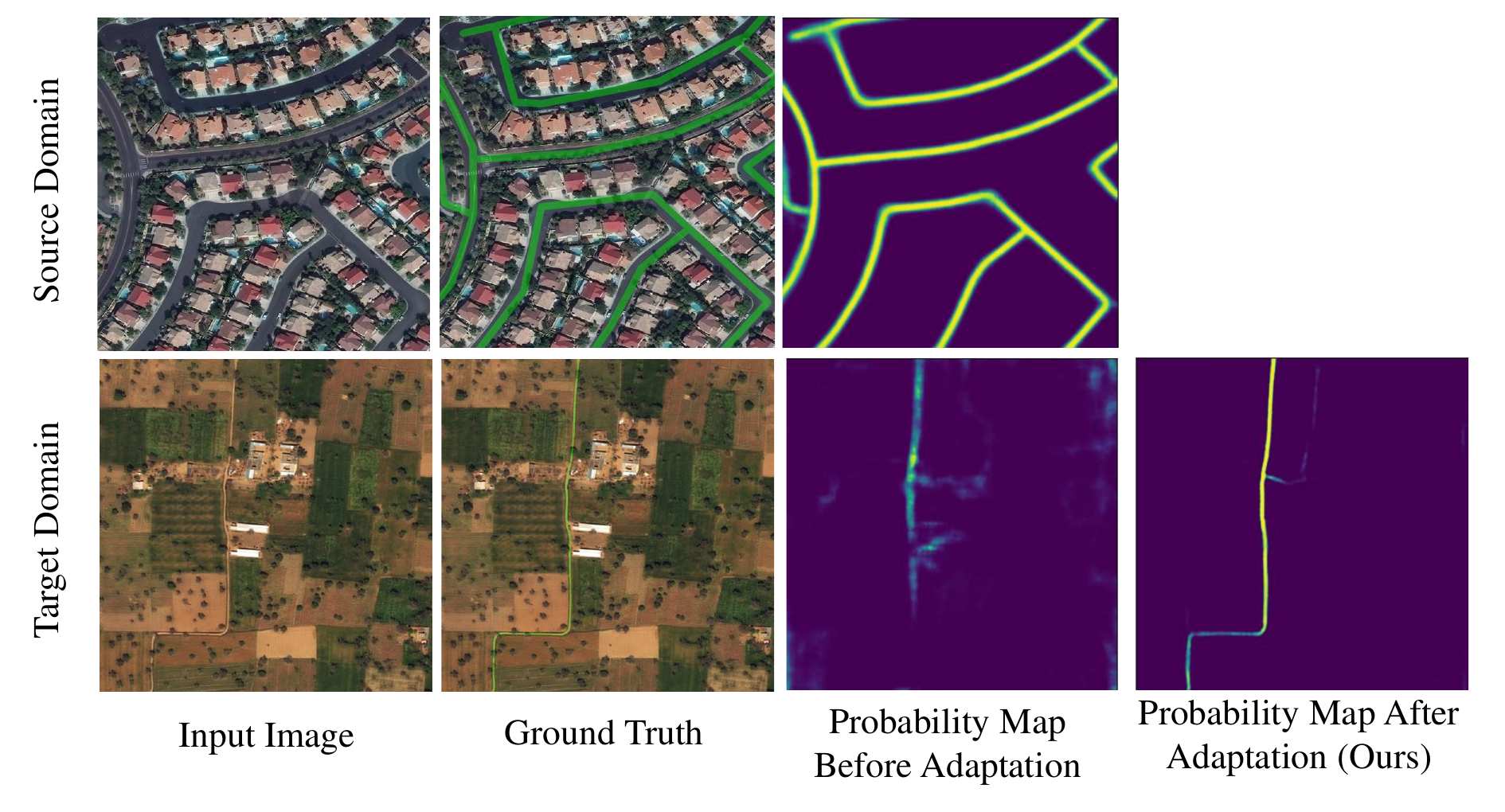}
    
    \caption{The source-trained model fails to generate accurate road segmentation for target domain images. Predicted roads in the target domain are broken and disconnected, and as indicated from the probability map, the probabilities slowly decrease while continuing in the direction of the road is too small to be labeled as a road. After the adaptation, we see better results with more spatial connectivity and completeness.  Best viewed in color. }
    \label{fig:intro_}
\end{figure*}

Recently, due to the availability of annotated datasets, such as DeepGlobe \cite{demir2018deepglobe} and SpaceNet \cite{van2018spacenet}, deep learning-based automatic road segmentation from satellite imagery has shown promising results \cite{etten2020city,he2020sat2graph,tan2020vecroad,mattyus2017deeproadmapper}. 
However, these models fail to segment roads accurately in unseen geographical regions \cite{lu2021cross, wang2020feature}.
This behavior is attributed to the \textit{domain shift} between the source (training) dataset and the unseen target (testing) dataset from another domain.
For satellite imagery, domain shift can occur due to the different image-capturing modalities, lighting conditions, geographical patterns, and the resolution difference between the source and target domains.

Compared to generic domain adaptation methods for semantic segmentation \cite{LSE_2020_Naseer,kim2020learning,pan2020unsupervised,zhang2020towards,wang2020differential,zhang2018fcan}, the domain adaptation for satellite road segmentation poses specific challenges. In the case of generic semantic segmentation, the structure of the scene (roads and objects are mostly in the lower part of the image and the sky is above) helps in domain adaptation.
Whereas, satellite/aerial imagery does not have such structure or geometry available \cite{iqbal2020weakly}. 
Road's appearance in satellite imagery is more dependent on the attributes of the geographical region; buildings on the edge of the road, natural environment, deserts, etc., and the adaptation performance usually suffers due to large and unstructured background area.
Additionally, based on the resolution of the satellite/aerial imagery, road width varies across regions and may only constitute a limited number of pixels making the road segmentation adaptation problem much more challenging.

To address the challenge of road segmentation domain adaptation, we propose a topology-based self-supervised adaptation strategy. 
Depending upon the construction, geography of the region, and satellite imagery resolution, road widths vary across the regions (source and target domains in our case) as illustrated in Figure  \ref{fig:intro_}. 
This further exacerbates the noisy nature of the pseudo-labels that are generated using the models trained on source data. 
Fig \ref{fig:intro_} also shows the deteriorated performance of the source model before adaptation. 
We observe that the road skeleton is well defined across domains, conceptually more domain invariant (i.e., does not depend on road width), and can help in topology preserving during domain adaptation. 
Hence, we incorporate road skeleton (center-line) prediction alongside road segmentation in a multi-task learning scenario, making the model more generalizable and using this auxiliary information to help in generating more refined pseudo labels during self-training. 
\textcolor{black}{
Note that the road skeleton is not only properly defined across domains and is conceptually more domain invariant (i.e., does not depend upon road width), it is helpful in preserving the topological concepts such as connectivity, continuity, and structure over the skeleton.
}

To enforce road connectivity, inspired by the classical line detection method \cite{bao2005canny}, we define a \textit{connectivity based pseudo-labels refinement} strategy (CBR). Using connectivity (neighborhood) information helps us improve the quality of pseudo-labels, which eventually increases the adaptation performance (Figure  \ref{fig:adap_dg} and Figure  \ref{fig:adap-mas}). To decrease the difference between source and target domain feature distribution, we perform adversarial learning at the encoder level. Note that we show that adversarial learning alone cannot align both domain features due to large data imbalance and is driven by the background (Table \ref{tab:res_adv}). 

To summarize, this work presents the following contributions.
We identify the major limitation of applying existing domain adaptation methods over the road-segmentation problem, i.e., these methods do not exploit the topological properties of the road. 
To overcome these limitations, we design a multi-task learning strategy of predicting the skeleton along with the road segmentation head, resulting in enriched features. 
Conformity loss between skeleton prediction and road segmentation is applied as a regularizer to guide the \textit{adaptation} over the unlabeled target domain.
We exploit the desired properties of the road such as connectivity and continuity, to clean the pseudo-labels (connectivity-based refinement of pseudo-labels)
resulting in better self-supervised learning/self-training. 
State-of-the-art domain adaption performance is achieved on benchmark road segmentation datasets.

\section{Related Work} 
In recent years, the accessibility of satellite imagery has improved multi-fold, resulting in the collection of large satellite imagery datasets.  
Deep learning has been applied to 
various problems including building/built-up regions segmentation \cite{bittner2017building,iqbal2020weakly,liu2021ct,chen2021dr}, 
destruction and change detection \cite{qiao2023weakly, ali2020destruction, nabiee2022hybrid, lv2022simple, lv2022spatial}, houses/structures counting \cite{liu2023cross, shakeel2019deep, lunga2020learning, zakria2021cross, nurkarim2023building}, and slums detection and mapping \cite{wurm2019semantic, rehman2022mapping} are explored. Despite the great success of deep learning-based approaches in remote sensing imagery, very little attention is devoted to domain adaptive road segmentation. In this section, we briefly review the road segmentation and adaptation methods related to the proposed approach.

\subsection{Road Segmentation} 
In recent years many deep learning-based models are proposed for semantic segmentation 
\cite{chen2018deeplab,badrinarayanan2015segnet,chen2014semantic,noh2015learning,ronneberger2015unet,zhao2017pyramid}. Mostly, these approaches are defined for ground road scenes or other generic scenes. 
Road segmentation in satellite and aerial images is a challenging problem due to large variations in background, shadows of the building, trees, etc. Modifying these generic segmentation methods, several approaches are designed for road extraction from aerial and satellite imagery \cite{etten2020city,batra2019improved}. The authors in \cite{batra2019improved} $\&$ \cite{mei2021coanet} tried to learn direction and connectivity alongside road segmentation respectively to preserve direction and connectivity information. Similarly, authors in \cite{zhou2018dlinknet} embedded dilated convolutions in \cite{chaurasia2017linknet} to better preserve edges and boundaries information. Although these approaches work reasonably well, they require large and precisely annotated road segmentation datasets, which are not trivial to collect.

Road skeleton segmentation can help in capturing the structure and topology of the roads which are important characteristics of road networks. 
The authors in \cite{wei2020simultaneous} used skeleton/center-line extraction to improve the broken links in road segmentation incrementally.
Whereas, \cite{alshaikhli2021simultaneous, cheng2017automatic} predict skeleton along road segmentation. \cite{alshaikhli2021simultaneous} reported only slight improvement in road segmentation when the skeleton prediction task was added. 
These methods require labeled data, do not attempt to perform domain alignment, and do not employ any structural or topological consistency, hence, producing deteriorated results when exposed to unseen target domain images. 
We explicitly enforce structural consistency by employing conformity loss, and for self-supervised domain adaptation, improve pseudo-label quality by performing connectivity-based refinement.

\subsection{Domain Adaptation}
Domain adaptation approaches \cite{tzeng2017adversarial,8578935,tsai2018learning,Benjdira_2019,ganin2016domainadversarial,iqbal2020mlsl,LSE_2020_Naseer,iqbal2020weakly,wang2021uncertainty} are widely used to minimize the domain shift between source and target domains. Adversarial learning is the most common approach to minimize the domain gap between source and target domain images segmentation either at input space \cite{hoffman2017cycada,long2015fully}, structured output space \cite{vu2019advent,clan_2019_CVPR,structure_2019_CVPR,dlow_2019_CVPR,tsai2018learning} or feature space representations \cite{chen2017no,iqbal2020weakly}. In recent years, many domain adaptation approaches based on self-training are proposed with state-of-the-art results for generic semantic segmentation \cite{zou2019confidence,zou2018unsupervised,LSE_2020_Naseer,iqbal2020mlsl,wang2021uncertainty,iqbal2022drsl}. The main idea of self-training based methods is, for each class, to select the most confident pixels as pseudo-labels and then finetune the source domain trained model using these pseudo-labels for target images \cite{LSE_2020_Naseer,zou2018unsupervised}.  These approaches exploit contextual information (such as vehicles will always be on the ground, and the sky will always be above) and since no such structure exists in satellite images,  these approaches fail to perform well on road segmentation problems. Similarly, the adversarial learning matches the global image features, or output probability distributions, which in the case of two classes like road segmentation, is overwhelmed by background (non-road areas).

Besides great progress in domain adaptation for semantic segmentation, the road segmentation adaptation is still an open challenge  \cite{zhang2021stagewise,shamsolmoali2020road,deng2019large,wang2020feature,lu2021cross}. The authors in \cite{zhang2021stagewise} presented two-stage adversarial learning, where the first stage aligns inter-domain global features and the second stage aligns features of intra-domain hard and easy examples using adversarial learning. Similarly, authors in  \cite{lu2021cross} tried to match the source and target features at global and local levels for better adaptation. However, these approaches defined for domain adaptation of road segmentation still fail to present road-specific constraints to preserve the characteristics and topology of the roads. Compared to these existing approaches, we explicitly try to preserve the road structure and connectivity by employing simultaneous road and skeleton segmentation and adaptation, structural conformity preservation loss, and connectivity-based pseudo-labels refinement. These problem-specific components significantly improve the performance compared to existing state-of-the-art methods.

\section{Methodology}
In this section, we provide details of {our proposed} unsupervised domain adaption (UDA) method for road segmentation.  
Our approach exploits the topological property of the road by performing connectivity-based pseudo-label refinement for improved pseudo-labels selection. 
During adaptation, the model is encouraged to learn domain invariant features through the road skeleton prediction task. 
Since the target domain is unlabeled, the skeleton prediction task is trained by applying consistency/conformity loss across the skeleton prediction and road segmentation head.
We use DLinkNet  \cite{zhou2018dlinknet}, as a base \textit{road segmentation method}. 
Due to dilated convolutions, high receptive field and skip connections, DLinkNet captures global, multi-scale information and contextual information, resulting in more accurate segmentation results (Table~\ref{tab:res-dg}).

\begin{figure*}[tb] 
    \centering
    \includegraphics[width=0.98\linewidth]{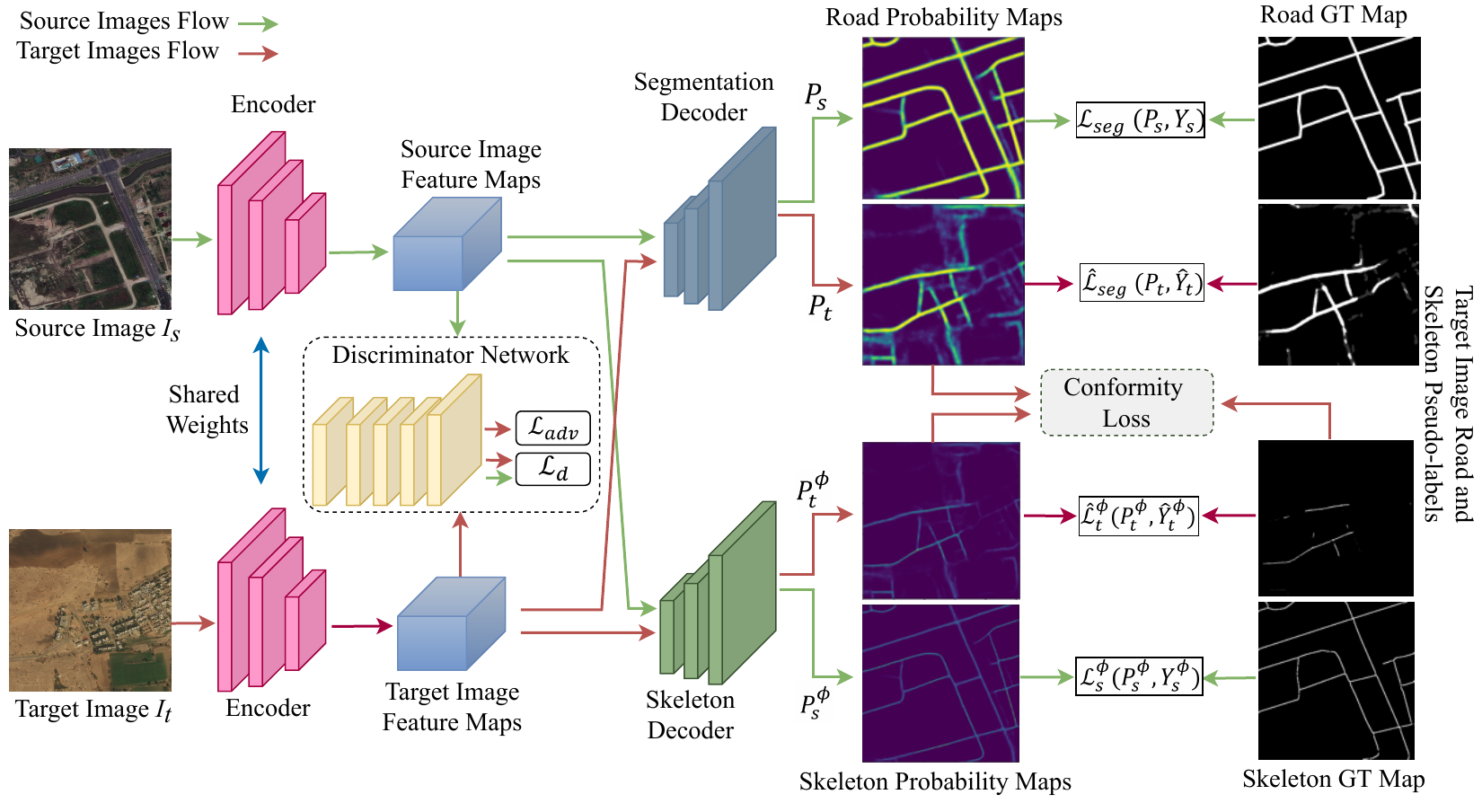}
    \caption{\textbf{Network Architecture:} We use a multi-head segmentation network to predict roads and skeleton segmentation with a shared encoder. Cross entropy loss for both road and skeleton is back-propagated for source and target domain images based on ground truth labels and pseudo-labels, respectively. A structural conformity loss is used to ensure the structural conformity between road and skeleton predictions \textcolor{black}{using the skeleton pseudo-labels}. Moreover, a discriminator is defined at the encoder to align the features of source and target domain images. During the evaluation, only the road segmentation head is used.}
    \label{fig:model-diagram}
\end{figure*}

\subsection{Preliminaries}
\label{sec:prelim}
\subsubsection{Problem Formulation}
Let \srcDomain~be the source domain consisting of labeled satellite image, \srcImg,~and the corresponding road segmentation labels, \srcLb. 
The objective of unsupervised domain adaptation is to generalize the model trained on \srcDomain~to the target domain \trgDomain,~where satellite images \trgImg~are available and ground-truth labels \trgLb~are not available. 
Without the loss of generality, we can assume \srcImg, \trgImg$\in \mathbb{R}^{H \times W\times 3}$ and \srcLb, \trgLb$\in \mathbb{Z}^{H \times W\times C}$,  where $H$ and $W$ are height and width of the image, and $C$ is the total number of classes a pixel can belong to. 
In a supervised learning setting for source domain \srcDomain, the objective is to learn segmentation model $g \in G : I \rightarrow Y$, by minimizing the cross-entropy loss function $\mathcal{L}_{seg}$, defined in \myEq{\ref{eqn:1-srcLoss_n}}.
\begin{align}
\scriptsize
 \mathcal{L}_{seg}(I_s, Y_s) =  - \frac{1}{H \times W}\sum_{h,w}^{H,W}\sum_{c}^{C} Y_{s}^{h,w,c}\log(P_s^{h,w,c})
\label{eqn:1-srcLoss_n}
\end{align}

let $g(I_s) = P_{s} \in \mathbb{R}^{H \times W\times C}$ 
is the segmentation probability map, where $P_{s}^{h,w,c}$ is probability of pixel $(h,w)$ belonging to class $c$.

\noindent\textbf{Pseudo-Label Selection: }
To overcome the non-availability of the $Y_{t}$ in the target domain, typical self-supervised adaptation methods perform pseudo-label selection from probable predictions \cite{iqbal2020mlsl,LSE_2020_Naseer,pan2020unsupervised, zou2018unsupervised}. 
Let $P_t =g(I_t)$ be the output probability volume for the target image $I_t$,
predicted by the source trained model, then pseudo-labels $\hat{Y_t}$ are computed by setting $\hat{Y_t}^{w,h,c}= \mathbbm{1}[P_t^{h,w,c}=\max(P_t^{h,w})]$.
Since these predictions might be noisy for \trgImg, low-confidence ones are masked out and not used during the back-propagation. 
\begin{align}
\begin{split}
&\mathcal{\hat{L}}_{seg}(I_t, \hat{Y}_t, m_t) =\phantom{d}\\
&- \frac{1}{H \times W}\sum_{h,w}^{H,W} m^{h,w}_t\sum_{c}^{C}\hat{Y}_t^{h,w,c}\log(P_t^{h,w,c})
\end{split}
\label{eqn:2-tarLoss_n}
\end{align}
where  $m^{h,w}_t \in \{0,1\}$ depending upon whether this pixel is in the pseudo-label set or not. Generally, $m^{h,w}_t =\mathbbm{1}[P_t^{h,w,k} \ge T]$, where $T$ is user defined threshold and $k = \argmax_c P_t^{h,w,c}$. 
The network is trained over the \srcLb, and pseudo labels $\hat{Y}_t$ (using $m$), which are updated after each round (set of convergence iterations, more details in Sec. \ref{sec:impl-details}). 

\subsubsection{Domain adaptation for road segmentation } 
In the case of road segmentation, Eq. \ref{eqn:1-srcLoss_n} and Eq. \ref{eqn:2-tarLoss_n} are reduced to binary cross entropy instead of categorical cross-entropy. 
This results in $P_{s}, P_{t} \in \mathbb{R}^{H \times W\times 1}$; representing the output probability map of whether a pixel belongs to the road class or not. 
The pseudo-labels are defined as $\hat{Y_t}^{h,w}= \mathbbm{1}[P_t^{h,w}  \ge (1-P_t^{h,w})]$. 
We backpropagate gradients for those pixels only where we are sure (e.g. have high predictive probability) that they either belong to the road or they do not belong road.
Therefore, mask $m_t$ is computed by checking two thresholds $m^{h,w}_t =\mathbbm{1}[P_t^{h,w} > T_r || (1-P_t^{h,w}) > T_b]$, where $T_r~ \text{and} ~T_b$ are thresholds for the road and background class (not belonging to the road), respectively.

\subsection{Multi-task self-supervised domain adaptation}

%


One of the components of the domain shift is the variations in  road width across domains, and how the boundary is delineated due to changes in the background (buildings on the edge of the road, natural environment, desert, etc.,).
On the other hand, the concept of the center-line or skeleton remains the same across domains. 
In addition, the skeleton in general embodies topological information like continuity and connectivity much better than what is captured by the road-surface segment.
Therefore, in this work, we employ skeleton prediction as an auxiliary task to improve road segmentation across domains (Figure  \ref{fig:model-diagram}).
The training labels for the skeleton prediction are obtained by skeletonizing true labels of the source domain \srcDomain~to the single pixel width using \cite{zhang1997fast}. 
The skeleton segmentation head is trained using Eq. \ref{eqn:3-srcLoss-sk},
%
\begin{align}
 \begin{split}
  \mathcal{L}_s^{{\phi}}(I_s, Y_s^{{\phi}}) =  - \frac{1}{H W} \sum_{h,w}^{H,W}[Y_s^{{\phi, (h,w)}}~\log(P_s^{{\phi}, (h,w)}) \\ + (1 - Y_s^{{\phi}, (h,w)})~ \log(1 - P_s^{{\phi}, (h,w)})],
 \end{split}
\label{eqn:3-srcLoss-sk}
\end{align}
where $\mathcal{L}_s^{{\phi}}(I_s, Y_s^{{\phi}})$ is the skeleton segmentation loss for source domain image $I_s$ with respective labels $Y_s^{{\phi}} \in \mathbb{R}^{H \times W}$ and predicted probability map $P_s^{{\phi}}\in \mathbb{R}^{H \times W}$. For target domain images, similar to road pseudo-labels, we define skeleton pseudo-labels based on the probabilities of the skeleton segmentation head and then use Eq. \ref{eqn:4-tarLoss-sk}. 
\begin{align}
\small
\begin{split}
  \mathcal{L}_t^{{\phi}}(I_t, \hat{Y}_t^{{\phi}}) =  - \frac{1}{HW}\sum^{H,W}_{w,h} m^{\phi, (h,w)}_t [\hat{Y}_t^{{\phi}, (h,w)}~\log(P_t^{{\phi}, (h,w)}) \\ + (1 - \hat{Y}_t^{{\phi}, (h,w)})~ \log(1 - P_t^{{\phi}, (h,w)})]
\end{split}
\label{eqn:4-tarLoss-sk}
\end{align}
%
where $m^{\phi, (h,w)}_t = \mathbbm{1}[P_t^{\phi, (h,w)} > T_r^{\phi} || (1-P_t^{\phi, (h,w)}) > T_b^{\phi}]$, and $T_r^{\phi}$ and $T_b^{\phi}$ are thresholds for the skeleton and background class, respectively. 
$m^{\phi, (h,w)}_t$ allows the selection of pseudo-labels, where either the model has predicted the road with high confidence or the background with high confidence.
$\hat{Y}_t^{{\phi}}\in \mathbb{R}^{H \times W}$ and $P_t^{{\phi}}\in \mathbb{R}^{H_t \times W_t}$ represent the skeleton pseudo-labels and predicted probability map for input target image $I_t$ respectively. 
\begin{figure*}[t]
    \centering
    \includegraphics[width=\linewidth]{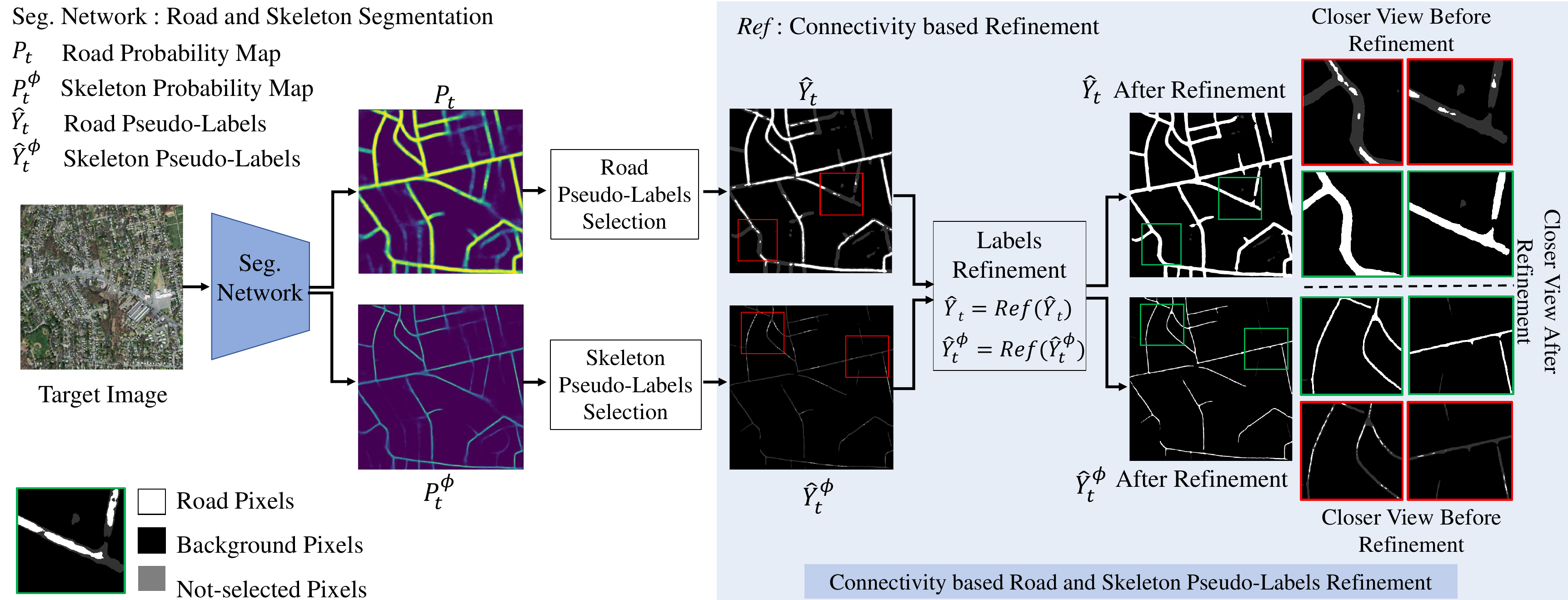}
    \caption{An illustration of the connectivity-based pseudo label refinement (CBR) approach for both road and skeleton pseudo labels. After applying CBR, the pseudo-labels are increased while reducing the number of `not-selected' pixels based on improving connectivity and completeness. 
    }
    \label{fig:conn-refinement}
    \vspace{-0.5cm}
\end{figure*}

\subsection{Connectivity-Based Pseudo Labels Refinement (CBR)}
\label{sec:cbr}
\begin{figure}[htb]
    \centering
    \includegraphics[width=0.8\linewidth]{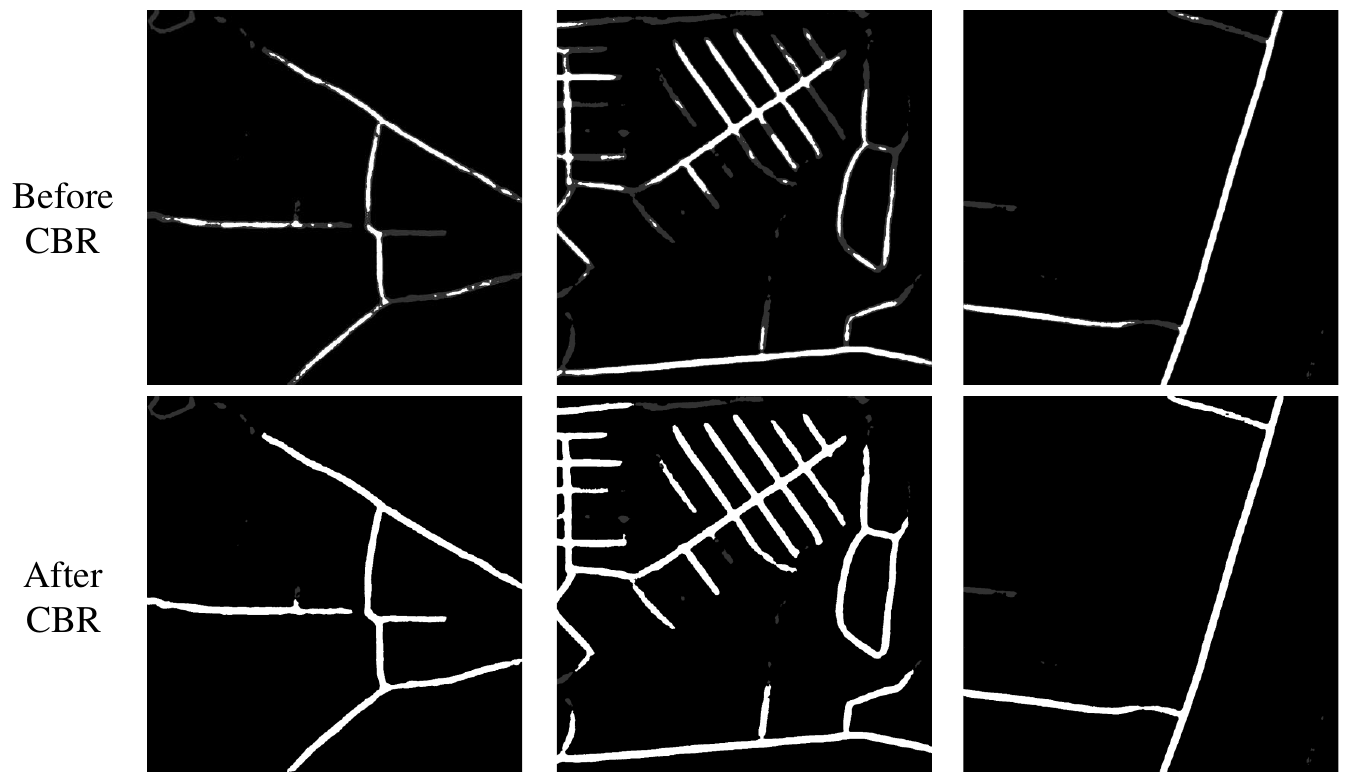}
    \caption{Pseudo-labels before and after connectivity-based refinement (CBR). The pseudo-labels are more complete and connected after CBR indicating their quality is improved significantly.
    }
    \label{fig:pl_prob}
\end{figure}

Effective self-training-based domain adaptation requires a reasonable number of good pseudo-labels for each class.
The confidence-based pseudo-label assignment and threshold-based masking (Sec.~\ref{sec:prelim}) results in highly imbalanced pseudo-labels. 
In the case of road segmentation, these pseudo-labels might result in broken segments or holes, thus the concept of continuity might not be enforced on the target domain as shown in Figure \ref{fig:pl_prob} (before CBR).

Note that the pseudo-label assignment and selection are based on the two thresholds, $T_r$ and $T_b$, where $T_r$ selects the road pixels, and $T_b$ is used for selecting non-road pixels. 
For ones having predicted probability in-between $T_r$ and $(1-T_b)$, we set $m^{h,w}_t=0$ (since we are not confident about their label, we call these pixels as 'not-selected' pixels), and they do not contribute to the loss (Eq. \ref{eqn:2-tarLoss_n}) and therefore are not used to update the weights of the model. 
Lowering $T_r$ will result in the background being labeled as roads and a high $T_r$ will result in disjoint road pieces. 
Therefore, we perform connectivity-based pseudo-label refinement, based on the two thresholds, $T_h(=T_r) $ and $T_l(>=1-T_b)$, to improve structure and connectivity in the road and skeleton pseudo-labels. 


Let $\mathcal{R} = \{\forall(h,w) | m^{h,w}_t =1~\&~P^{h,w}_t>T_{h}\}$ be set of selected pixels assigned the pseudo-label road. Let $\mathcal{M}_t = \{\forall(h,w) | m^{h,w}_t =0\}$ be set of all `not-selected' pixels. 
Any pixel in $\mathcal{M}_t$ that is connected to the road pseudo-labeled pixels in  $\mathcal{R}$ is selected as the road pseudo-label. For all, $(h,w) \in \{ (P_{t}^{h,w} > T_l)~\&~(P_{t}^{h,w} < T_h)~\&~\zeta((h,w), R) = \text{True}\}$, we set $m^{h,w}_t=\mathbbm{1}$ and $\hat{Y}_t^{h,w}=1$. 
Where $\zeta((h,w), R)$ is the Boolean function that uses the 8-neighborhood to identify if $(h,w)$ is connected to any pixel in $\mathcal{R}$. 
\textcolor{black}{
This process is recursive and repeated until all pixels in $\mathcal{R}$ are traversed and checked for connectedness with the existing road labels as well as the newly assigned road labels. }
A similar approach is used for skeleton pseudo-labels selection with $T_h^{\phi}$ and $T_l^{\phi}$ as upper and lower thresholds respectively.  
This process of pseudo-label selection and connectivity-based refinement is shown in Figure  \ref{fig:conn-refinement}. These connectivity-based refined pseudo-labels are used during self-supervised domain adaptation.

\subsection{Structure Conformity Loss}
\label{sec:consistency}

To further guide the adaptation process, we enforce spatial consistency between thin skeleton prediction and much wider (road) segment predictions by defining a structural conformity loss (Eq. \ref{eqn:5-consLoss}) between the output of two heads. 
More specifically, we apply an $L_2$-distance loss between road and skeleton probabilities. 
The skeleton head tries to predict a single pixel center-line.
During UDA for the source images, we compute the difference between the prediction of the skeleton and road segmentation, only for the pixels which are labeled  as a skeleton by ground truth.
Similarly, in the case of images from the target domain, we compute the difference between the prediction of the skeleton and road segmentation, only for the pixels which are \textit{pseudo-labeled}  as a skeleton \textcolor{black}{ as shown in Figure \ref{fig:model-diagram} (red arrow originating from skeleton pseudo-labels to conformity loss)}. Combined conformity loss $\mathcal{L}_{cl}$ is given below. 
\begin{align}
\footnotesize
\begin{split}
  \mathcal{L}_{cl} =  \sum_{H,W} [Y_s^{{\phi}}= 1] (P_s - P_s^{{\phi}})^2 +  \sum_{H,W} [\hat{Y}_t^{{\phi}}= 1] (P_t - P_t^{{\phi}})^2.
\end{split}
\label{eqn:5-consLoss}
\end{align}
Proposed conformity loss enforces that the thinning of the road-surface results in the same skeleton as the one predicted by the skeleton segmentation head. 
An overview of conformity loss for target images is shown in Figure  \ref{fig:model-diagram}.  

%

\subsection{Adversarial Learning based Feature Alignment}

Finally, we employ adversarial learning to align the source and target domain image features. 
We define a discriminator network $D$ at the encoder level of the segmentation network, and try to minimize the gap between source and target domain features as shown in Figure  \ref{fig:model-diagram}. 
The discriminator network ($\mathcal{L}_{d}$, Eq. \ref{eqn:6-disc}) is trained using the cross-entropy loss for source and target domain images. 
Similarly, the adversarial loss ($\mathcal{L}_{adv}$, Eq. \ref{eqn:7-adv}) for the target domain is used to update the encoder of the segmentation network. Let the features at the output of the encoder are denoted by $F_e \in \mathbb{R}^{h_f\times w_f\times C_f}$, with $h_f, w_f$ and $C_f$ as height, width, and depth of the feature map respectively, the $\mathcal{L}_{d}$ and $\mathcal{L}_{adv}$ are defined as,
\begin{align}
\begin{split}
    \mathcal{L}_d (F_e) =  - \frac{1}{h_f \times w_f}\sum_{h_f,w_f} [y_d~\log(D({F_e}))\\ + (1-y_d)~\log(1 - D({F_e})))]
 \label{eqn:6-disc}
\end{split}
\end{align}
\begin{align}
    \mathcal{L}_{adv}(F_{e_t}) =  - \frac{1}{h_f \times w_f} \sum_{h_f,w_f} \log(D({F_{e_t}}))
\label{eqn:7-adv}
\end{align}
\textcolor{black}{$F_{e_t}$ represents the encoder feature maps of the target images while $y_d$ shows the domain (source, target) label to train the discriminator and generate an adversarial loss.}



\noindent\textbf{Total Loss Function: }
During domain adaptation, we simultaneously train the road and skeleton segmentation head using the labeled source data and the pseudo-labeled target data. 
The composite loss function (Eq. \ref{eqn:4-composit-seg-loss}) is the summation of the individual losses, 
Eq. \ref{eqn:1-srcLoss_n},  \ref{eqn:2-tarLoss_n},  \ref{eqn:3-srcLoss-sk}, and \ref{eqn:4-tarLoss-sk}.
\begin{align}
\begin{split}
  \mathcal{L}_{comp} =  \mathcal{L}_{seg}(I_s, Y_s) + \mathcal{L}_s^{{\phi}}(I_s, Y_s^{{\phi}})\\  + \hat{\mathcal{L}}_{seg}(I_t, \hat{Y_t}) + \hat{\mathcal{L}}_t^{{\phi}}(I_t, \hat{Y}_t^{{\phi}}) 
\end{split}
\label{eqn:4-composit-seg-loss}
\end{align}
Finally, the total loss (Eq. \ref{eqn:8-total}) is the summation of composite loss (Eq. \ref{eqn:4-composit-seg-loss}), conformity loss (Eq. \ref{eqn:5-consLoss}) and adversarial loss (Eq. \ref{eqn:7-adv}).    
\begin{align}
    \mathcal{L}_{total} = \mathcal{L}_{comp} + \beta \mathcal{L}_{cl} + \lambda_{adv}\mathcal{L}_{adv}
\label{eqn:8-total}
\end{align}
where, hyper-parameters $\beta$, and $\lambda_{adv}$ are chosen empirically (Sec. \ref{sec:impl-details} and \ref{sec:ablation-exp}).


\begin{figure}[t]
    \centering
    \includegraphics[width=1.0\linewidth]{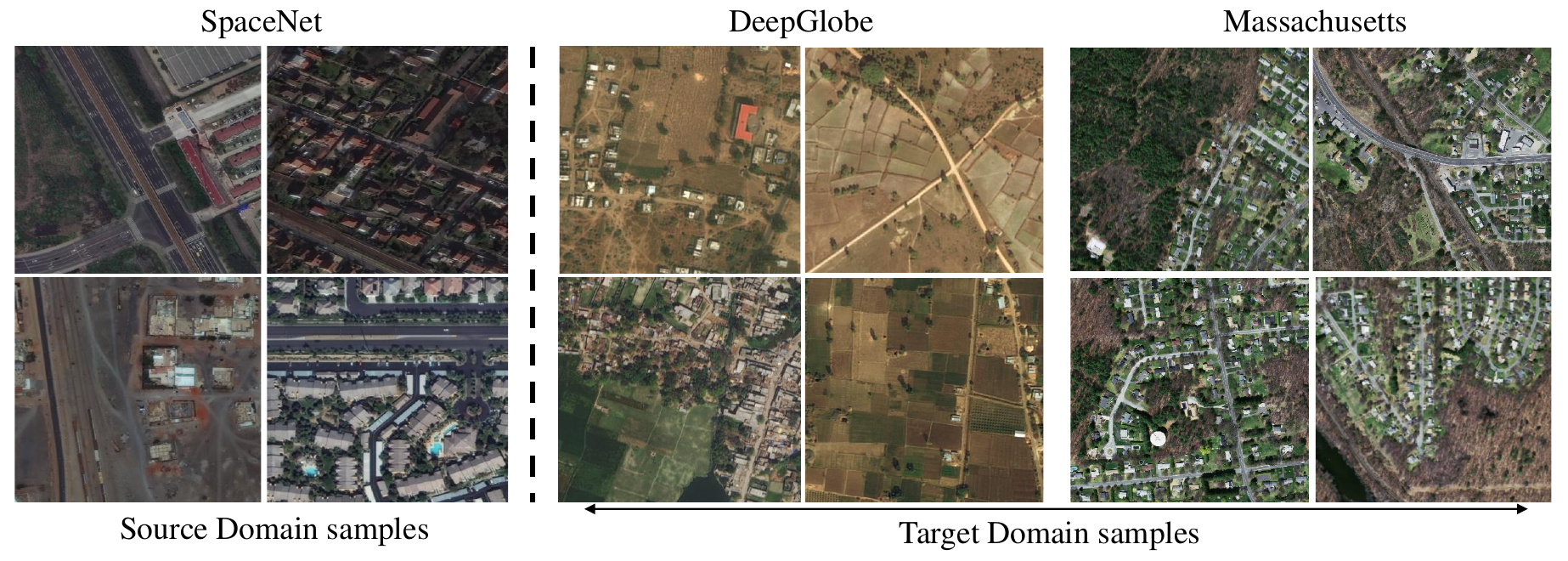}
    \caption{Example images showing the visual, infrastructural, and scale differences in Source and Target domains.}
    \label{fig:data}
\end{figure}

\section{Experiments}

\subsection{Experimental Setup}

\noindent\textbf{Datasets: }
\label{sec:dataset}
For evaluation, we use SpaceNet \cite{van2018spacenet}, DeepGlobe \cite{demir2018deepglobe} and Massachusetts \cite{mnih2013machine} as benchmark road segmentation datasets. 
The SpaceNet, DeepGlobe, and Massachusetts datasets are different with respect to colorization, geographies, spatial resolution, texture, illuminations, and road structures. A few sample images from datasets are shown in Fig \ref{fig:data}. \textbf{SpaceNet} dataset consists of annotated satellite imagery for road segmentation for 4 different cities, i.e., Paris, Las Vegas, Shanghai, and Khartoum. 
There are a total of 2780 image tiles with a spatial resolution of 1300$\times$1300 pixels covering 30\textit{cm}$/$pixel. However, labels are available for a subset of 2548 image tiles. SpaceNet covers around 3000 \textit{sq.km} of area
and around 8000\textit{km} annotated roads. 
\textbf{DeepGlobe} consists of 6226 labeled image tiles of size 1024$\times$1024, at 50~\textit{cm}$/$pixel. 
\textcolor{black}{The DeepGlobe dataset is obtained by capturing satellite imagery over Indonesia, Thailand, and India \cite{demir2018deepglobe}. For available labeled images, we have followed the training and validation split defined by \cite{batra2019improved}}.
\noindent\textbf{Massachusetts} is an aerial imagery road segmentation dataset covering 100~\textit{cm} per-pixel. There are a total of 1171 labeled images of size 1500 $\times$ 1500 with a defined split of 1108 as training, 14 as validation, and 49 as testing images, respectively.

\noindent \textbf{Model Architecture: }
Our proposed method is mainly composed of segmentation and discriminator networks. The segmentation network (also known as the generator network) is based on DLinkNet \cite{zhou2018dlinknet} with ResNet-34 \cite{he2016deep} as the backbone feature extractor.
We follow the same architecture as the original DLinkNet \cite{zhou2018dlinknet} for road segmentation to have a fair comparison.
A separate decoder-head is attached to the encoder for the skeleton prediction, as shown in Figure  \ref{fig:model-diagram}. 
The discriminator network is inspired by \cite{tsai2018learning} and composed of four convolution layers each with the filter size 4$\times$4 followed by LeakyRelu \cite{xu2020reluplex} activation except the last layer, where sigmoid is used as the activation.


\noindent \textbf{Implementation and Training Details: }
\label{sec:impl-details}
The proposed domain adaptation approach is implemented using the PyTorch framework with a single Core-i5 machine with 11GB of GPU memory. From all the images, we randomly crop $1024 \times 1024$ image patch for processing, and the batch size is set to two. 
The initial learning rates are set to $2e^{-4}$ and $1e^{-4}$ for self-supervised adaptation and adversarial learning, respectively.
The adversarial loss is scaled with $\lambda_{adv} = 0.01$ to reduce the effect of large gradients. 
Similarly, the conformity loss is scaled with $\beta = 0.1$.
The sensitivity analysis of $\beta$ and $\lambda_{adv}$ are detailed in Sec.~\ref{sec:ablation-sens}. 
To select road pseudo-labels and mask $m^{h,w}$, thresholds are set as, $T_r=T_h = 0.9$,   $T_b = 0.3$ and  $T_l = 1 - T_b$. 
Similarly, for skeleton pseudo-labels and mask $\hat{m}^{h,w}$, the $T_r^{\phi}=T_h^{\phi}$ and $T_b^{\phi}$ are set to be 0.5 and 0.9 respectively, where $T_l^{\phi} = 1-T_b^{\phi}$. 

The DLinkNet \cite{zhou2018dlinknet} road segmentation network is trained on the source domain. 
\textcolor{black}{During adaptation, we follow an iterative process, i.e., generate pseudo-labels for the whole target dataset (by fixing the segmentation network) and then train that segmentation model over pseudo-labeled target data alongside labeled source data. 
Following the terminology of \cite{fleuret2021uncertainty,iqbal2020mlsl,munir2021ssal,iqbal2022fogadapt}, we define this iterative process of pseudo-labels generation and model training as rounds.} In the initial round (round-0), the segmentation model (trained on source data) is further trained for skeleton segmentation alongside road segmentation using the labeled source data. 
After round-0, before the start of the next rounds (round-1 and round-2),  pseudo-labels for target images are generated using the updated model and then used for retraining that model for two epochs alongside fully labeled source data. 


\subsection{Experimental Results}
\label{sec:exp-res}
For all the experiments, SpaceNet is used as the source domain dataset while DeepGlobe and Massachusetts are used as target domain datasets. 
Following \cite{iqbal2020weakly,zhang2021stagewise}, intersection over union (IoU) \& F1-Score are used as evaluation metrics for road segmentation. 
To better assess the connectivity and connectivity of the segmented roads, we follow \cite{he2020sat2graph, tan2020vecroad, bastani2018roadtracer} and also report APLS (Average Path Length Similarity).
For a more fair evaluation, we compare the proposed approach with generic DA methods (\cite{zou2018unsupervised, iqbal2020mlsl, zou2019confidence, wang2021uncertainty,vu2019advent,chen2019domain_maxsquare} ) as well as road-specific DA methods including TGN\cite{lu2021cross}, GOAL\cite{lu2021cross}, RoadDA\cite{zhang2021stagewise} and DOER\cite{lu_doer_tip}.

\begin{table}[t]
\centering
\caption{Comparison between the proposed  approach and existing approaches for SpaceNet to DeepGlobe and SpaceNet to Massachusetts adaptation. \textcolor{black}{\textbf{Oracle} results (upper-bound) are obtained when the segmentation model is trained over the target domain images using the ground truth labels \cite{iqbal2020mlsl, iqbal2020weakly}.
Backbone: Backbone architecture used in the segmentation network. Adv: Adversarial adaptation, ST: Self-training based adaptation, Adv+: Adversarial learning with any other adaptation approach. }
} 
\resizebox{\textwidth}{!}{
\scriptsize
\begin{tabular}{c|c|c|ccc|ccc}
\hline
& & &\multicolumn{3}{c}{SpaceNet $\rightarrow$  DeepGlobe}&\multicolumn{3}{|c}{SpaceNet $\rightarrow$  Massachusetts} \\
\hline
Method & Backbone & Approach & IoU & F1-Score & APLS & IoU & F1-Score & APLS\\
\hline

Source-only \cite{lu2021cross}& \multirow{6}{*}{ResNet-101} & - & 17.7 & 30.1 &10.3 & - & - & -\\
RoadDA \cite{zhang2021stagewise}&  & Adv & 23.3 & 36.3 &14.4 & - & - & -\\ 
MinEnt \cite{vu2019advent}&  & ST & 27.5 & 43.2 & - & - & - & -\\
MaxSquare \cite{chen2019domain_maxsquare}&  & ST & 27.7 & 43.4 &- & - & - & -\\ 
TGN \cite{lu2021cross}&  & Adv & 27.8 & 43.6 &- & - & - & -\\
GOAL \cite{lu2021cross}&  & Adv+ & 32.1 & 48.6 &- & - & - & -\\ 
DOER \cite{lu_doer_tip}&  & Adv+ & 32.8 & 49.4 &- & - & - & - \\  \hline
Source-only \cite{chen2018deeplab}& \multirow{3}{*}{ResNet-101} & - &  29.2 & 40.0 &24.6 & 24.0 & 38.8 & 5.70\\
CRST-MRKLD  \cite{zou2019confidence}&  & ST &  33.0 & 49.6 &37.6 & 29.5 & 45.6 & 47.2\\ 
Model-Uncertainty\cite{fleuret2021uncertainty}&  & ST &  33.4 & 50.1 &- & 29.8 & 46.0 & -\\  \hline
Source-only \cite{wu2019wider}& \multirow{3}{*}{ResNet-38} & - &  37.2 & 54.0 & 29.8 & 26.9 & 42.4 & 12.5\\
CBST \cite{zou2018unsupervised}&  & ST &  39.0 & 56.1 &39.6 & 28.4 & 44.3 & 49.1\\
MLSL-SISC \cite{iqbal2020mlsl}&  & ST & 39.6 &56.5 &40.4 & 30.1 & 46.1 & 42.1\\ \hline
Source-only \cite{zhou2018dlinknet}& \multirow{2}{*}{ResNet-34} & - & 35.2 & 47.4 &39.2 & 23.0 & 37.4 & 21.0\\
Ours&  & ST+Adv & \textbf{46.2} & \textbf{63.2} &\textbf{50.2} & \textbf{34.0} & \textbf{50.7} & \textbf{50.8}\\
\hline
Oracle& ResNet-34 & - & 61.2 & 75.0 &64.9 & 60.0 & 73.5 & 67.9\\
\hline
\end{tabular}
}
\label{tab:res-dg}
\end{table}


\noindent \textbf{SpaceNet $\rightarrow$  DeepGlobe:}
Table~\ref{tab:res-dg} presents a comparative analysis of the proposed road adaptation approach with different source models and existing state-of-the-art methods.
Compared to the Source-only (DLinkNet) model, the proposed approach improves the IoU, F1-score, and APLS by a margin of 11\%, 15.8\%, and 11\%, respectively.
\textcolor{black}{Compared to TGN\cite{lu2021cross}, GOAL \cite{lu2021cross}, and DOER \cite{lu_doer_tip}, recent adversarial learning-based methods for domain adaptation of road segmentation, the proposed approach outperforms by significant margin of 18.4\%, 14.1\%, and 13.4\% in IoU and 19.6\%, 14.6\%, and 13.8\% in F1-Score, respectively.} 
\textcolor{black}{Further, compared to another recent approach for road segmentation adaptation \cite{zhang2021stagewise}, the proposed approach shows an improvement by a significant margin of 22.9\% in IoU, 26.9\% in F1-score, and 35.8\% in APLS, respectively. }
Similarly, compared to ResNet-38 \cite{wu2019wider} based self-training approaches CBST \cite{zou2018unsupervised} and MLSL-SISC \cite{iqbal2020mlsl}, the proposed approach outperforms by a minimum margin of 6.6\%, 6.7\%, and 9.8\% in IoU, F1-Score, and APLS, respectively. 
We also compare our results with DeepLab-v2 \cite{chen2018deeplab} based self-training methods CRST \cite{zou2019confidence} and Model-Uncertainty \cite{fleuret2021uncertainty}, where the proposed approach outperforms with a minimum of margins of 12.8\% and 13.1\% in IoU and F1-Score, respectively. 
The considerable improvement in all the evaluation metrics, APLS, F1-Score, and IoU indicates the improvement of segmented roads in target images after adaptation. The gain in APLS indicates the segmented roads after adaptation are more continuous and connected. 
Compared to competing methods, the proposed approach results are close to the Oracle results. \textcolor{black}{The \textbf{Oracle} results, also known as upper-bound, are obtained when the segmentation model is trained over the target domain images using their original ground truth labels as defined by \cite{tsai2018learning,iqbal2020mlsl,hoffman2017cycada,iqbal2020weakly,tsai2020learning}.
The \textbf{Oracle} model serves as a benchmark or upper bound for any segmentation model to compare the performance of the adaptation approach.
 }
\begin{figure}[!htb]
    \centering
    \includegraphics[width=\columnwidth]{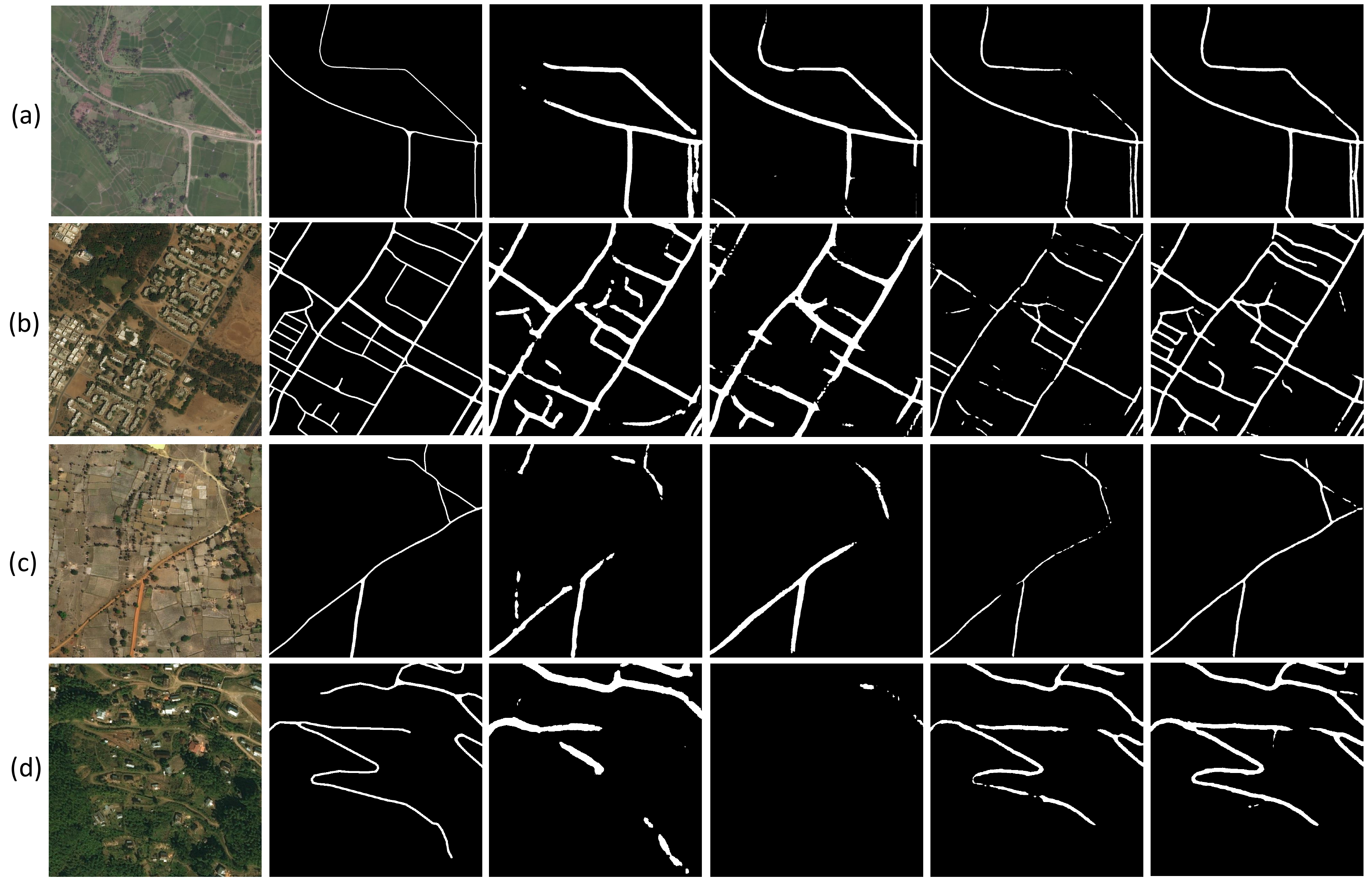}
    \scriptsize
    \begin{tabular}{P{0.001cm}P{1.6cm}P{1.5cm}P{1.65cm}P{1.3cm}P{1.7cm}P{1.3cm}}
    \scriptsize
    &Target Image & Ground Truth & Source-only (DLinkNet \cite{zhou2018dlinknet}) &CRST \cite{zou2018unsupervised} & Ours* (Without CBR) & Ours
    \end{tabular}
    \vspace{-0.1cm}
    \caption{SpaceNet $\rightarrow$ DeepGlobe adaptation: Compared to the Source-only model (DLinkNet\cite{zhou2018dlinknet}) and baseline CRST \cite{zou2019confidence}, the proposed approach produces better segmentation results with comparatively better completeness and connectivity. ``Ours*: Without CBR" and ``Ours: With CBR". 
    }
    \label{fig:adap_dg}
\end{figure}

\begin{figure}[!htb]
    \centering
    \includegraphics[width=\linewidth]{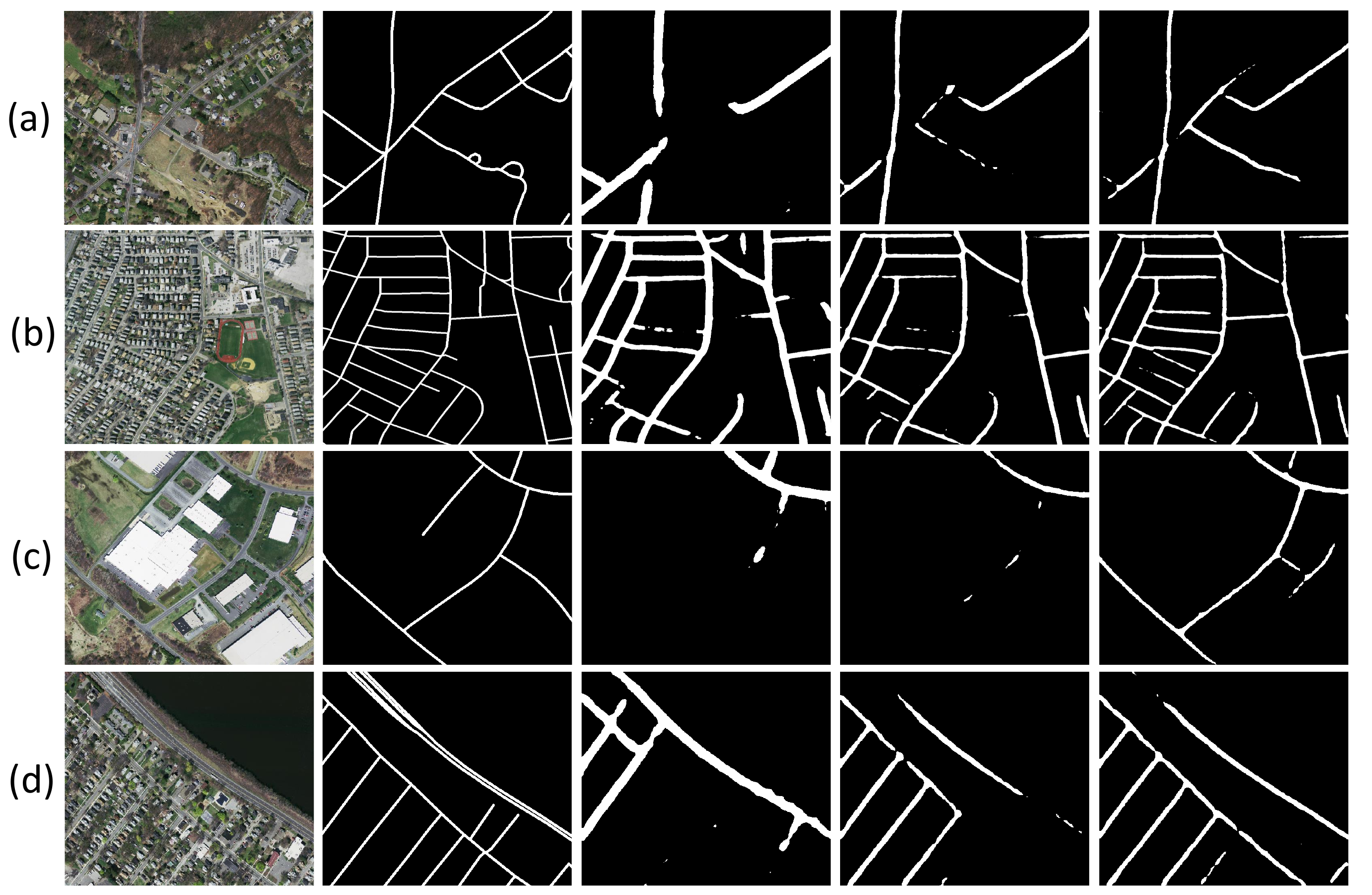}
    \scriptsize
    \begin{tabular}{P{0.005cm}P{1.95cm}P{1.96cm}P{1.96cm}P{1.96cm}P{1.95cm}}
    \scriptsize
    &Target Image & Ground Truth & Source-only (DLinkNet \cite{zhou2018dlinknet})  &  Ours* (Without CBR) & Ours
    \end{tabular}
    \caption{From SpaceNet to Massachusetts adaptation. Compared to the source model (DLinkNet \cite{zhou2018dlinknet}), our approach produces better segmentation results.
    }
    \label{fig:adap-mas}
\end{figure}

\noindent \textbf{SpaceNet $\rightarrow$  Massachusetts:}
As visible in Figure  \ref{fig:data}, the Massachusetts dataset, similar to the SpaceNet dataset, consists of images from the region where roads, buildings, and other structures appear to be appropriately planned. 
The key challenges for adaptation from the SpaceNet to Massachusetts are the difference in spatial resolution ($100 cm^2$ compared to SpaceNet's $30 cm^2$ per pixel), visual appearance, and the roads to be thin when compared to the source domain. 

The proposed approach outperforms the competing methods with a minimum margin of 3.9\% in IoU, 4.6\% in F1-Score, and 1.6\% in APLS (Table \ref{tab:res-dg}). More specifically, we gain 11.0\% in IoU, 13.3\% in F1-Score, and 29.7\% in APLS over the Source-only model (DLinkNet). Similarly, compared to ResNet-38 \cite{wu2019wider}, CBST \cite{zou2018unsupervised} and MLSL-SISC \cite{iqbal2020mlsl}, we gain 7.1\%, 5.6\% and 3.9\% in IoU,  8.3\%, 6.4\% and 4.6\% in F1-Score, and 38.3\%, 1.7\% and 8.7\% in APLS, respectively. 
Compared to DeepLab-v2 \cite{chen2018deeplab}, CRTS \cite{zou2019confidence} and Model-Uncertainty \cite{fleuret2021uncertainty} by a minimum margin of 4.2\%, 4.7\% and 3.6\% in IoU, F1-Score and APLS, respectively. 



\subsection{Qualitative Results.}
Figure \ref{fig:adap_dg} \&  \ref{fig:adap-mas} shows the qualitative results. 
It can be seen that the Source-only segmentation model (DLinkNet [39]) produces a lot of false negatives (missing roads or segments) as well as false positives (e.g., predicting thick/wider roads).
\textcolor{black}{This behavior is attributed to the resolution and road-width differences between the source and target domain images. Specifically, SpaceNet-3 (source domain dataset), has 0.3m/pixel and wider roads while the DeepGlobe and Massachusetts (target datasets) have comparatively low resolution  (0.5m/pixel and 1m/pixel, respectively) and thinner roads.}
The proposed approach tries to mitigate both these issues by increasing the true road segmentation and avoiding the non-road areas.
Our proposed connectivity-based pseudo-labels refinement (CBR) further improves the completeness and connectivity of the road segmentation as shown in Figure \ref{fig:adap_dg} \&  \ref{fig:adap-mas}, ``Ours*: Without CBR" and ``Ours: With CBR". 


\subsection{Ablation Experiments}
\label{sec:ablation-exp}
To evaluate the effect of different components, we perform domain adaptation by removing different components (Table \ref{tab:ablation-dg}). 
Note that for the CL and CBR, the round-0 is performed where the skeleton head is trained along with fine-tuning of the segmentation head over the source dataset (Sec. \ref{sec:impl-details}). 

\begin{table}[h]
\centering
\caption{Component-wise analysis of our approach, for two sources, one with skeleton and another without it. SK: Skeleton segmentation, ST: Self-training, FSA: Feature space adaptation, CL: Conformity loss, CBR: Connectivity-based pseudo-labels refinement.}

\resizebox{0.8\columnwidth}{!}{

\begin{tabular}{c|c|c|c|c|c|c|c|c} 
\hline
&\multicolumn{5}{c|}{Components} & \multicolumn{3}{c}{SpaceNet $\rightarrow$ DeepGlobe} \\
\hline
 Approach &SK &ST &FSA  &CL &CBR &IoU & F1-Score & APLS \\
\hline
Source-only&-&- &-  &-  &- & 35.2 & 47.4 &35.3 \\ \hline
\multirow{3}{*}{Ours} &- &- &\checkmark  &- &-  & 37.8 & 54.9 &-\\
  &- &\checkmark &-  &- &-  & 37.4 & 53.4 &38.4\\
&- &\checkmark &\checkmark  &- &-  & 40.3 & 57.4  &41.8\\ \hline
Source-only&\checkmark &- &-  &-  &- & 36.1 & 53.1  &39.2\\ \hline
\multirow{5}{*}{Ours}&\checkmark &\checkmark &-  &- &-  &  39.7 & 56.8  &38.1 \\
&\checkmark &- &\checkmark  &- &-  &  39.2 & 56.3  &- \\
&\checkmark &\checkmark &-   &\checkmark &-  &  42.7 & 59.9   &41.1\\
&\checkmark &\checkmark &\checkmark &\checkmark &-  &  44.2 & 61.3  &45.4\\
&\checkmark&\checkmark &\checkmark  &\checkmark &\checkmark  &  \textbf{46.2} & \textbf{63.2}  &\textbf{50.3}\\
\hline
\end{tabular}
}

\label{tab:ablation-dg}
\vspace{-0.4cm}
\end{table}


\subsubsection{Effect of skeleton segmentation and Conformity Loss:}
The skeleton (center-line) segmentation helps in preserving the road's topology and is a key component of the proposed road-segmentation-adaptation approach. 
Due to the multi-task learning, introducing a skeleton segmentation head improves the accuracy of the road segmentation (source) model from 35.2 IoU to 36.1 IoU, as shown in Table \ref{tab:ablation-dg} (Source-only Results with and without skeleton). \textcolor{black}{Similarly, the APLS score is improved from 35.3 to 39.2 indicating improved connectivity of the segmented roads.
Secondly, identifying the skeleton also allows us to apply the conformity loss between the road and skeleton heads alongside self-training, resulting in a performance improvement of 3.0\% in IOU, 3.1\% in F1-score, and 3.0\% in APLS for SpaceNet to DeepGlobe adaptation.
Similarly, we apply CBR over the skeleton head to refine the skeleton pseudo-labels, making both the self-training and conformity loss more effective.
These additional controls improve our results from 40.3\% to 46.2\% in IoU, 57.4\% to 63.2\% in F1-Score, and from 41.8\% to 50.3\% in APLS score. 
With only a skeleton segmentation head and conformity loss (with self-training), we see 7.5\% points improvement in IoU over the initial source model. }
For details look at Table \ref{tab:ablation-dg}.

\subsubsection{Connectivity based Pseudo-labels Refinement (CBR)}

\begin{figure*}[!htb]
    \centering
    \includegraphics[width=\linewidth]{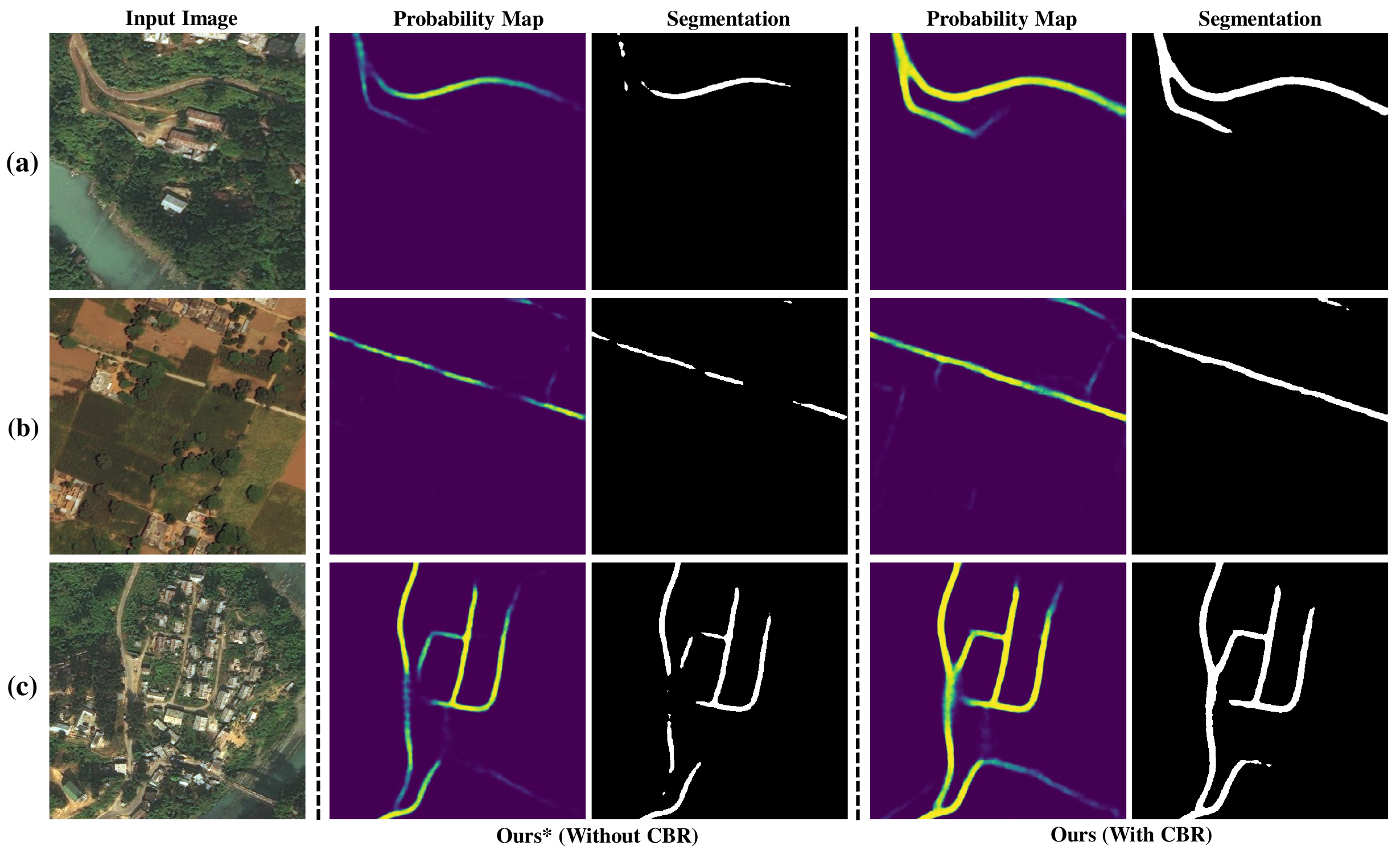}
    \scriptsize
    \caption{\textcolor{black}{SpaceNet to DeepGlobe adaptation. The output probabilities and resulting segmentation are significantly improved after CBR. The roads are more continuous and connected compared to those without CBR.
    }}
    \label{fig:cbr_prob}
\end{figure*}

To understand the effect of CBR, we analyze the quality of pseudo-labels in terms of IoU, F1-Score, and APLS with ground truth (only used for evaluation here) with and without CBR (Table \ref{tab:res_connect_pl}). 
We see a considerable improvement in the quality of pseudo-labels (Figure \ref{fig:pl_prob}), resulting in a more accurate prediction of roads in the target domain after adaptation (Table \ref{tab:ablation-dg}). Specifically, the roads segmented have more connectivity and continuity (Figure \ref{fig:adap_dg} \& \ref{fig:adap-mas}), thus resulting in high APLS. 
\textcolor{black}{
To further highlight the effect of CBR, we show the output probability maps and their respective segmentation masks (after thresholding) without and with CBR in Figure \ref{fig:cbr_prob}. It can be seen that in all the cases, the probabilities of road prediction improve significantly with the introduction of CBR. Thus, results in continuous and connected roads. }

\begin{table}[h!]
\caption{Pseudo-labels quality before and after CBR.}
\centering


\resizebox{0.75\columnwidth}{!}{
\begin{tabular}{c|ccc|ccc} 
\hline
\multicolumn{7}{c}{SpaceNet $\rightarrow$ DeepGlobe} \\
\hline \hline
 & \multicolumn{3}{c|}{Round-1} & \multicolumn{3}{c}{Round-2} \\ \hline
Process & IoU &F1-Score & APLS &IoU &F1-Score & APLS \\ \hline

Before Refinement &  34.7 & 51.5 & 27.0 &  44.9 & 61.9 & 45.3 \\
After Refinement &  \textbf{37.0} & \textbf{54.0} & \textbf{35.7} &  \textbf{49.3} & \textbf{66.0} & \textbf{48.2}\\
\hline
\end{tabular}
}
\label{tab:res_connect_pl}
\end{table}

\begin{table}
\caption{Adversarial learning-based adaptation.}
\centering

\resizebox{0.5\columnwidth}{!}{
\begin{tabular}{c|c|c} 

\hline
\multicolumn{3}{c}{SpaceNet $\rightarrow$ DeepGlobe} \\
\hline
Method & IoU &F1-Score \\ \hline
Source-only (DLinkNet) \cite{zhou2018dlinknet} &  35.2 & 47.4 \\
Output Space Adaptation (OSA) \cite{tsai2018learning} & 37.3 & 54.2 \\
Feature Space Adaptation (FSA) \cite{chen2017no}&  37.8 & 54.9 \\
\hline
\end{tabular}
}
\label{tab:res_adv}

\end{table}

\subsubsection{Adversarial Learning based Feature Space Adaptation} 

\textcolor{black}{In general  adversarial learning-based domain adaptation is a difficult task due to high sensitivity and slow convergence. However, many approaches have been applied to adapt the segmentation models either at structured output space \cite{vu2019advent,clan_2019_CVPR,structure_2019_CVPR,dlow_2019_CVPR,tsai2018learning}, or latent features space \cite{chen2017no,iqbal2020weakly}.
For road driving scenes like, GTA \cite{Richter_2016_ECCV} and Cityscapes \cite{Cordts2016Cityscapes} datasets, the output space adversarial domain is mostly applied because there exists a naturally defined structure, i.e., the sky always on the top side of the image, roads always on down rows of the image, etc. However, no such structure is defined in the case of remote sensing imagery. Secondly, in the case of road segmentation in satellite imagery, adversarial learning is overwhelmed by large backgrounds (non-road areas). In this work, we explored both feature space and output space adversarial learning-based adaptation for satellite imagery.  
}

Table \ref{tab:res_adv} shows that feature space adversarial learning adaptation (FSA) performs  better compared to output space adaptation (OSA),
but still, the improvement due to adversarial learning alone is only comparable to simple self-training based adaptation.
However, when FSA is combined with the self-training, the road segmentation performance is improved by 2.9\% in IoU for SpaceNet to DeepGlobe adaptation (Table \ref{tab:ablation-dg} 4th row). Hence, performing adversarial learning along with other components results in better feature alignment, as visible in the last row of Table \ref{tab:ablation-dg}.

\subsubsection{Adapting from DeepGlobe to SpaceNet}
To further validate the effectiveness of the proposed approach, we performed experiments in the opposite direction (DeepGlobe (source) to SpaceNet (target)). Our proposed
method improves IoU from 25.6\% (before DA) to 29.8\%.

\subsubsection{Sensitivity Analysis of Hyperparameters}
\label{sec:ablation-sens}

In this section, we provide the sensitivity analysis of the hyper-parameters  $\lambda_{adv}$, $\beta$, and pseudo-labels selection thresholds $T_h$ and $T_l$, used during our adaptation approach.

\noindent\textbf{Sensitivity of $\beta$ (Conformity loss scaling factor) :}
We investigate the effect of conformity loss weight $\beta$ (Eq. \ref{eqn:8-total}) on the adaptation performance of the proposed approach (Table~\ref{tab:beta}). The experiments in Table~\ref{tab:beta} show that the conformity loss in general is robust to the scaling factor and improves the performance even when no scaling factor, i.e., $\beta = 1.0$ is used. However, better performance is reported for $\beta = 0.1$.


\begin{table}[!h]
\caption{Effect of conformity loss weight $\beta$ on the adaptation process.}
\centering


\resizebox{0.45\columnwidth}{!}{
\begin{tabular}{c|cccc} 
\hline
\multicolumn{5}{c}{SpaceNet $\rightarrow$ DeepGlobe} \\
\hline \hline
$\beta$ & 0.001 &0.01 &0.1 &1.0 \\ \hline
IoU &  43.9 & 45.3 &  46.2 & 44.2 \\
F1-Score &  60.9 & 62.3 &  63.2 & 61.3 \\
\hline
\end{tabular}
}
\label{tab:beta}
\end{table}

\noindent\textbf{Effect of $\lambda_{adv}$ :}
In Table \ref{tab:lambda}, we show the effect of adversarial loss scaling factor $\lambda_{adv}$ along with self-supervised domain adaptation. The adversarial loss helps to improve the adaptation (Table \ref{tab:ablation-dg}), however, is comparatively more sensitive to loss weight as shown in Table \ref{tab:lambda}. The higher value of $\lambda_{adv}$ propagates large gradients to the network, while a very small value of $\lambda_{adv}$ may not help the adaptation process significantly. A similar observation was reported by \cite{tsai2018learning} as well.


\begin{table}[!htb]
\caption{Effect of $\lambda_{adv}$, adversarial loss scaling factor in Eq. \ref{eqn:8-total}. The results reported are for adversarial learning combined with self-supervised learning only.}
\centering


\resizebox{0.375\columnwidth}{!}{
\begin{tabular}{c|ccc} 
\hline
\multicolumn{4}{c}{SpaceNet $\rightarrow$ DeepGlobe} \\
\hline \hline
$\lambda_{adv}$ & 0.001 &0.01 & 0.1 \\ \hline
IoU &  37.7 & 40.3 & 37.3 \\
F1-Score & 54.1 & 57.4 & 51.3 \\
\hline
\end{tabular}
}
\label{tab:lambda}
\end{table}



\noindent\textbf{Effect of $T_h$ and $T_l$ :}
We also analyze the effect of upper and lower thresholds for connectivity-based refinement (CBR) of pseudo-labels. Table \ref{tab:thresholds}, shows the adaptation performance for different combinations of upper and lower thresholds. It can be seen that reducing the upper threshold to 0.8 deteriorates the overall performance. This is attributed to the false positives (low precision) selected as pseudo-labels, eventually decreasing the IoU as well as F1-Score significantly. The lower threshold for CBR is comparatively less sensitive to overall adaptation performance. However, reducing $T_l$ to 0.5 and less affects the adaptation performance (Table \ref{tab:thresholds}). 


\begin{table}[!htb]
\caption{Effect of $T_h$ and $T_l$ for road pseudo-labels selection.}
\centering


\resizebox{0.3\columnwidth}{!}{
\begin{tabular}{cc|cc}
\hline
\multicolumn{4}{c}{SpaceNet $\rightarrow$ DeepGlobe} \\
\hline \hline
$T_h$ &$T_l$ &IoU &F1-Score \\ \hline

0.9 & 0.7 &  46.2 & 63.2 \\
0.9 & 0.5 &  44.8 & 62.3 \\
0.8 & 0.6 &  43.2 & 60.4 \\
\hline
\end{tabular}
}
\label{tab:thresholds}
\end{table}


\subsubsection{Analysis of Backbone Architectures}
\label{sec:ablation-backbone}
\textcolor{black}{
We have performed additional experiments with our proposed road segmentation adaptation approach by changing the backbone networks. 
} 


\begin{table}[h!]
\caption{\textcolor{black}{Comparison of ResNet-34 and large backbone architectures (ResNet-101, SWIN) used with the DLinkNet segmentation model to evaluate the segmentation and adaptation performance of the proposed approach. It can be observed that despite computation limitations, our adaptation strategy improves models’ accuracy considerably over the target domain.}}
\centering


\resizebox{0.75\columnwidth}{!}{
\begin{tabular}{c|c|c|ccc} 
\hline
Approach &Backbone & Image Size &IoU &F1-Score & APLS \\ \hline

Source-only & \multirow{2}{*}{ResNet-34} & \multirow{2}{*}{ 1024 x 1024} & 35.2 & 47.4 & 35.3 \\
Ours &  &  & 46.2 &  63.2 & \textbf{50.3} \\ \hline

Source-only & \multirow{2}{*}{ResNet-101} & \multirow{2}{*}{512 x 512} & 38.0 & 55.1 & 32.6 \\
Ours  &  &  & 45.94 &  63.0 & 47.9 \\ \hline

Source-only& \multirow{2}{*}{Swin-B} & \multirow{2}{*}{512 x 512} & 39.0 & 56.1 & 37.7 \\
Ours  &  &  & \textbf{46.7} &  \textbf{63.7} & 49.2 \\ \hline

\end{tabular}
}
\label{tab:ablation-back}
\end{table}


\textcolor{black}{
By using larger backbones, e.g., Swin-B and ResNet-101, the Source Model’s accuracy improves by 3.8\% and 2.8\%  compared to ResNet-34 based DLinkNet, respectively. However, these larger backbones bases segmentation models still suffer from Domain Shift. As indicated in Table \ref{tab:ablation-back}, our adaptation strategy improves models’ accuracy considerably over the target domain (7.94\% and 7.7\% in IoU and 15.3\% and 11.5\% in APLS scores, for ResNet-101 and Swin-B based segmentation models, respectively).}

\textcolor{black}{
Please note that, due to limited computational resources, we were unable to train and adapt using the large backbone models (ResNet-101 and Swin-B) with large image sizes. Therefore we have reduced the  size of images to  512 x 512 for  ResNet-101 and Swin-B to best utilize our resources, as indicated in Table \ref{tab:ablation-back}. The results in Table \ref{tab:ablation-back} reinforce that regardless of the backbone, road-segmentation (Source-only) models suffer from the domain shift, and our adaptation strategy results in performance improvement (ours).
We believe that having large memory GPUs, and increasing the image sizes (as used in the case of ResNet-34) will further improve the results with ResNet-101 and Swin-B backbone architectures. Thus, we argue that the proposed approach improves the performance, is complementary, and can be used with any backbone architecture.
}

\section{Conclusions}
In this paper, we tackle the challenging problem of overcoming the domain shift for road segmentation algorithms. 
The proposed self-supervised domain adaptation method exploits the topology of the road by having a structural conformity loss across the skeleton of the road and the road surface itself. 
The quality of the pseudo-labels for self-training is improved by introducing connectivity-based pseudo-labels refinement. To help the adaptation process, source, and target features are aligned using discriminator-based adversarial learning.
Our thorough experimental results on three different datasets on multiple evaluation metrics, multiple backbone architectures, detailed ablation studies, and comparison with several competitive baselines validate the proposed ideas and framework.



{\small
\bibliography{egbib}
}

\end{document}